\documentclass[conference,letterpaper]{IEEEtran}
\usepackage{fancyhdr}
\usepackage{graphicx}  

\usepackage{amsmath}   

\usepackage{fancyhdr}

\begin{document}
\title{Node discovery in a networked organization}

\author{\IEEEauthorblockN{Yoshiharu Maeno}
\IEEEauthorblockA{Social Design Group\\
Bunkyo-ku, Tokyo 112-0011, Japan.\\
maeno.yoshiharu@socialdesigngroup.com}
}


%


\maketitle
\thispagestyle{plain}

\fancypagestyle{plain}{
\fancyhf{}	
\fancyfoot[L]{978-1-4244-2794-9/09/\$25.00~\copyright~2009 IEEE} 
\fancyfoot[C]{}
\fancyfoot[R]{SMC 2009}
\renewcommand{\headrulewidth}{0pt}
\renewcommand{\footrulewidth}{0pt}
}

\pagestyle{fancy}{
\fancyhf{} 
\fancyfoot[R]{SMC 2009}}
\renewcommand{\headrulewidth}{0pt}
\renewcommand{\footrulewidth}{0pt}

\begin{abstract}
In this paper, I present a method to solve a node discovery problem in a networked organization. Covert nodes refer to the nodes which are not observable directly. They affect social interactions, but do not appear in the surveillance logs which record the participants of the social interactions. Discovering the covert nodes is defined as identifying the suspicious logs where the covert nodes would appear if the covert nodes became overt. A mathematical model is developed for the maximal likelihood estimation of the network behind the social interactions and for the identification of the suspicious logs. Precision, recall, and F measure characteristics are demonstrated with the dataset generated from a real organization and the computationally synthesized datasets. The performance is close to the theoretical limit for any covert nodes in the networks of any topologies and sizes if the ratio of the number of observation to the number of possible communication patterns is large.
\end{abstract}


%
\IEEEpeerreviewmaketitle



 



\section{Introduction}
\label{Introduction}

Covert nodes in a networked organization refer to persons who affect social interactions (communications among the nodes and resulting collaborative activities), but do not appear in the surveillance logs which record the participants of the social interactions. They are not observable directly. Discovering the covert nodes is defined as identifying the suspicious surveillance logs where the covert nodes would appear if the covert nodes became overt. This problem is called a node discovery problem.

Where do we encounter such a problem? Globally networked clandestine organizations such as terrorists, criminals, or drug smugglers are great threat to the civilized societies. Terrorism attacks cause great economic, social and environmental damage. Active non-routine responses to the attacks are necessary as well as the damage recovery management. The short-term target of the responses is the arrest of the perpetrators. The long-term target of the responses is identifying and dismantling the covert organizational foundation which raises, encourages, and helps the perpetrators. The threat will be mitigated and eliminated by discovering covert leaders and critical conspirators of the clandestine organizations. The difficulty of such discovery lies in the limited capability of surveillance. Information on the leaders and critical conspirators are missing because it is usually hidden by the organization intentionally.

In this paper, I present a method to solve the node discovery problem. The method infers the network topology and probability parameters behind the social interactions (by use of the maximal likelihood estimation), applies an anomaly detection technique to the surveillance logs, and identifies the suspicious surveillance logs. \ref{Method} presents the method. \ref{Test} introduces the dataset generated from a real organization and the computationally synthesized datasets for performance tests. \ref{Performance} demonstrates the precision, recall, and F measure characteristics with the datasets.

\section{Related Work}
\label{Related Work}

The social network analysis is a study of social structures made of nodes which are linked by one or more specific types of relationship. Examples of the relationship are influence transmission in communication, or presence of trust in collaboration. Studies in complex networks \cite{Bar99}, \cite{Wat98}, \cite{Erd59}, WWW search and analysis \cite{Cha07}, \cite{Ada01}, and machine learning of latent variables \cite{Sil06}, \cite{Sin04} are major related research topics.

Research interests have been moving from describing organizational nature to discovering unknown phenomena. A link discovery predicts the existence of an unknown link between two nodes from the information on the known attributes of the nodes and the known links \cite{Cla08}, \cite{Get05}, \cite{Tas03}. The link discovery techniques are combined with domain-specific heuristics. The collaboration between scientists can be predicted from the published co-authorship \cite{Lib04}. The friendship between people is inferred from the information available on their web pages \cite{Ada03}. Discovery of a network structure \cite{Rab08}, \cite{New07}, \cite{Pal05} and detection of an anomaly in a network \cite{Sil09} are also relevant related research topics.

A node discovery predicts the existence of an unknown node around the known nodes from the information on the collective behavior of the network. Related works in the node discovery is limited. Heuristic method for node discovery is proposed in \cite{Mae07}. The method applies clustering algorithm \cite{Zak08}, \cite{Has01} to the nodes in a network, and detects the node which inter-connects clusters at the border of a cluster in clustered networks. The method is applied to analyze the covert social network foundation behind the terrorism disasters \cite{Mae09}.

\section{Method}
\label{Method}

\subsection{Observation}
\label{Observation}

A node and a link in a social network are a person and a relationship resulting in influence transmission between persons. The symbols $n_{j} \ (j=0,1,\cdots)$ represent the nodes. Some nodes are overt (observable), but the others are covert (unobservable). $\mbox{\boldmath{$O$}}$ denote a set of the whole overt nodes $\{n_{0}, n_{1}, \cdots, n_{N-1} \}$. Its cardinality is $N=|\mbox{\boldmath{$O$}}|$. $\mbox{\boldmath{$C$}} = \overline{\mbox{\boldmath{$O$}}}$ denotes a set of the whole covert nodes $\{n_{N},n_{N+1}, \cdots \}$. The symbol $\delta_{i} \ (0 \leq i < D)$ represent an individual communication pattern (and a resulting collaborative activity) among the persons. It is a set of nodes, $\delta_{i} \in \mbox{\boldmath{$O$}} \cap \mbox{\boldmath{$C$}}$. The unobservability of the covert nodes does not affect the communication pattern. For example, the members of a communication pattern are those who join an online community.

An observation $d_{i}$ in surveillance logs is a set of the overt nodes in a communication pattern $\delta_{i}$. It is given by eq.(\ref{Dataset2}). The number of data is $D$.
\begin{equation}
d_{i} = \delta_{i} \cap \mbox{\boldmath{$O$}} \ (0 \leq i < D).
\label{Dataset2}
\end{equation}

$\{ d_{i} \}$ denotes the observation dataset. Note that neither an individual node nor a single link can be observed directly, but a group of nodes can be observed as a communication pattern. $\{ d_{i} \}$ can be expressed by a 2-dimensional $D \times N$ matrix of binary variables $\mbox{\boldmath{$d$}}$. The presence or absence of the node $n_{j}$ in the data $d_{i}$ is indicated by the elements in eq.(\ref{BasketVector}).
\begin{equation}
\mbox{\boldmath{$d$}}_{ij} = \left \{ \begin{array}{ll}
                    1 & \mbox{\ \ if $n_{j} \in d_{i}$} \\
                    0 & \mbox{\ \ otherwise}
                \end{array}
         \right . (0 \leq i < D, \ 0 \leq j < N).
\label{BasketVector}
\end{equation}

\subsection{Maximal Likelihood Estimator Network}
\label{Likelihood}

A parametric form is defined to describe the network topology and the influence transmission over the network. The influence transmission governs the possible communication patterns $\{\delta_{i}\}$ which result in the observation dataset $\{ d_{i} \}$. The probability where the influence transmits from an initiating node $n_{j}$ to a responder node $n_{k}$ is $r_{jk}$. The influence transmits to multiple responders independently in parallel. It is similar to the degree of collaboration probability in trust modeling \cite{Lav07}. The constraints are $0 \leq r_{jk}$ and $\sum_{k \neq j} r_{jk} \leq 1$. The quantity $f_{j}$ is the probability where the node $n_{j}$ becomes an initiator. The constraints are $0 \leq f_{j}$ and $\sum_{j=0}^{N-1} f_{j} = 1$. These parameters are defined for the whole nodes in a social network (both the nodes in $\mbox{\boldmath{$O$}}$ and $\mbox{\boldmath{$C$}}$).

A single symbol $\mbox{\boldmath{$\theta$}}$ represent both of the parameters $r_{jk}$ and $f_{j}$ for the nodes in $\mbox{\boldmath{$O$}}$. $\mbox{\boldmath{$\theta$}}$ is the target variable, the value of which needs to be inferred from the observation dataset. The logarithmic likelihood function \cite{Has01} is defined by eq.(\ref{Likelihood1}). The quantity $p(\{ d_{i} \}|\mbox{\boldmath{$\theta$}})$ denote the probability where the observation dataset $\{ d_{i} \}$ realizes under a given \mbox{\boldmath{$\theta$}}.
\begin{equation}
L(\mbox{\boldmath{$\theta$}}) = \log( p(\{ d_{i} \}|\mbox{\boldmath{$\theta$}})).
\label{Likelihood1}
\end{equation}

The individual observations are assumed to be independent. eq.(\ref{Likelihood1}) becomes eq.(\ref{Likelihood2}).
\begin{equation}
L(\mbox{\boldmath{$\theta$}}) = \log( \prod_{i=0}^{D-1} p(d_{i}|\mbox{\boldmath{$\theta$}})) = \sum_{i=0}^{D-1} \log( p(d_{i}|\mbox{\boldmath{$\theta$}})).
\label{Likelihood2}
\end{equation}

The quantity $q_{i|jk}$ in eq.(\ref{fijk1}) is the probability where the presence or absence of the node $n_{k}$ as a responder to the stimulating node $n_{j}$ coincides with the observation $d_{i}$. 
\begin{equation}
q_{i|jk} = \left \{ \begin{array}{ll}
                    r_{jk} & \mbox{\ \ if $\mbox{\boldmath{$d$}}_{ik} = 1$ for given $i$ and $j$} \\
                    1-r_{jk} & \mbox{\ \ otherwise}
                \end{array}
         \right ..
\label{fijk1}
\end{equation}

eq.(\ref{fijk1}) is equivalent to eq.(\ref{fijk2}) since the value of $d_{ik}$ is either $0$ or $1$.
\begin{equation}
q_{i|jk} = \mbox{\boldmath{$d$}}_{ik} r_{jk} + (1-\mbox{\boldmath{$d$}}_{ik})(1-r_{jk}).
\label{fijk2}
\end{equation}

The probability $p(\{ d_{i} \}|\mbox{\boldmath{$\theta$}})$ in eq.(\ref{Likelihood2}) is expressed by eq.(\ref{Prob1}). The operator $\wedge$ means logical AND.
\begin{equation}
p(d_{i}|\mbox{\boldmath{$\theta$}}) = \sum_{j=0}^{N-1} \mbox{\boldmath{$d$}}_{ij} f_{j} \prod_{0 \leq k < N \ \wedge \ k \neq j} q_{i|jk}.
\label{Prob1}
\end{equation}

The logarithmic likelihood function takes an explicit formula in eq.(\ref{Likelihood3}). The case $k=j$ in multiplication ($\prod_{k}$) is included since $d_{ik}^{2}=d_{ik}$ always holds.
\begin{equation}
L(\mbox{\boldmath{$\theta$}}) = \sum_{i=0}^{D-1} \log ( \sum_{j=0}^{N-1} \mbox{\boldmath{$d$}}_{ij} f_{j} \prod_{k=0}^{N-1} \{ 1-\mbox{\boldmath{$d$}}_{ik}+(2\mbox{\boldmath{$d$}}_{ik}-1)r_{jk} \} ).
\label{Likelihood3}
\end{equation}

The maximal likelihood estimator $\hat{\mbox{\boldmath{$\theta$}}}$ is obtained by solving eq.(\ref{Estimator}).
\begin{equation}
\hat{\mbox{\boldmath{$\theta$}}} = \arg \max_{\mbox{\boldmath{$\theta$}}} L(\mbox{\boldmath{$\theta$}}).
\label{Estimator}
\end{equation}

A simple incremental optimization technique (hill climbing method) is employed to solve eq.(\ref{Estimator}). Simulated annealing method \cite{Has01} can be employed to strengthen the search ability and to avoid sub-optimal solutions. These methods search more optimal parameter values around the present values and update them as in eq.(\ref{HillClimb}) until the values converge.
\begin{equation}
\left \{ \begin{array}{l}
                    r_{jk} \rightarrow r_{jk} + \Delta r_{jk} \\
                    f_{j} \rightarrow f_{j} + \Delta f_{j}
         \end{array}
\right . (0 \leq j, k < N) .
\label{HillClimb}
\end{equation}

The update $\Delta r_{nm}$ and $\Delta f_{n}$ should be in the direction of the maximal ascend of the likelihood function. It is indicated by the multiplication of the derivatives and the updates in eq.(\ref{deltaL}).
\begin{equation}
\Delta L(\mbox{\boldmath{$\theta$}}) = \sum_{n,m=0}^{N-1} \frac{\partial L(\mbox{\boldmath{$\theta$}})}{\partial r_{nm}} \Delta r_{nm} + \sum_{n=0}^{N-1} \frac{\partial L(\mbox{\boldmath{$\theta$}})}{\partial f_{n}} \Delta f_{n}.
\label{deltaL}
\end{equation}

Individual derivatives in eq.(\ref{deltaL}) are calculated by eq.(\ref{Partiallikelihood1}), and eq.(\ref{Partiallikelihood2}).
\begin{eqnarray}
\frac{\partial L(\mbox{\boldmath{$\theta$}})}{\partial r_{nm}} = \sum_{i=0}^{D-1} [ f_{n} \mbox{\boldmath{$d$}}_{in} (2\mbox{\boldmath{$d$}}_{im}-1) \prod_{k \neq m} \{ 1-d_{ik} +(2\mbox{\boldmath{$d$}}_{ik} \nonumber \\
-1) r_{nk} \} \div \sum_{j=0}^{N-1} \mbox{\boldmath{$d$}}_{ij} f_{j} \prod_{k=0}^{N-1} \{ 1-\mbox{\boldmath{$d$}}_{ik}+(2\mbox{\boldmath{$d$}}_{ik}-1)r_{jk} \} ].
\label{Partiallikelihood1}
\end{eqnarray}
\begin{eqnarray}
\frac{\partial L(\mbox{\boldmath{$\theta$}})}{\partial f_{n}} = \sum_{i=0}^{D-1} [ \mbox{\boldmath{$d$}}_{in} \prod_{k=0}^{N-1} \{ 1-\mbox{\boldmath{$d$}}_{ik}+(2\mbox{\boldmath{$d$}}_{ik}-1)r_{nk} \} \nonumber \\
\div \sum_{j=0}^{N-1} \mbox{\boldmath{$d$}}_{ij} f_{j} \prod_{k=0}^{N-1} \{ 1-\mbox{\boldmath{$d$}}_{ik}+(2\mbox{\boldmath{$d$}}_{ik}-1)r_{jk} \} ].
\label{Partiallikelihood2}
\end{eqnarray}

\subsection{Node Discovery - Anomaly Detection}

Suspiciousness of the observation data $d_{i}$ is evaluated by eq.(\ref{Ranking1}). Suspiciousness means the likeliness where the covert node would appear in the data if it became overt. Larger value means more suspicious data. 
\begin{equation}
s(d_{i}) = \frac{1}{p(d_{i}|\hat{\mbox{\boldmath{$\theta$}}})}.
\label{Ranking1}
\end{equation}

Ranking of the observation data can be calculated from the value of eq.(\ref{Ranking1}). The $i$-th most suspicious data is given by $d_{\sigma(i)}$ in eq.(\ref{Ranking2}). Suspiciousness $s(d_{\sigma(i)})$ is larger than $s(d_{\sigma(i')})$ for any $i<i'$. 
\begin{equation}
\sigma(i) = \arg \max_{m \neq \sigma(n) \ {{\rm for}} \ ^{\forall}n<i} s(d_{m}) \ (1 \leq i \leq D).
\label{Ranking2}
\end{equation}

\section{Test Dataset}
\label{Test}

\subsection{Network Model}
\label{Network}

Two classes of network models are employed to generate communication test dataset. The first class is a real organization. The second class is a mathematical model having several adjustable parameters.

\begin{figure}
\begin{center}
\includegraphics[scale=0.31,angle=90]{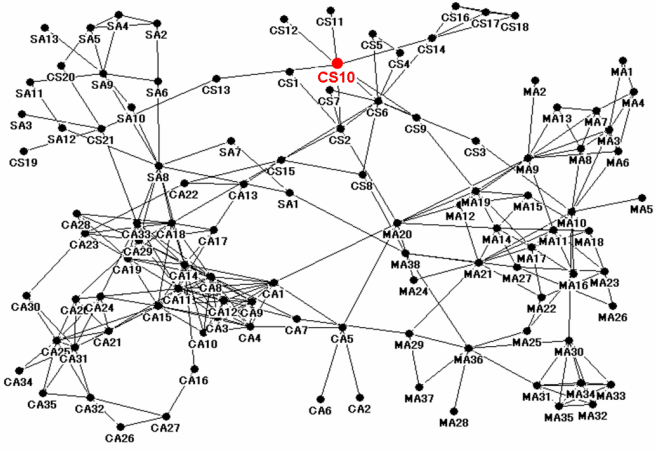}
\end{center}
\caption{Network model (A) representing global mujahedin (Jihad fighters) organization \cite{Sag04}. The model consists of $107$ nodes, and 4 regional sub-networks. The sub-networks represent Central Staffs (CS), Core Arabs (CA), Maghreb Arabs (MA), and Southeast Asians (SA). The node $n_{{\rm CS10}}$, which is indicated by a red circle, is believed to be the founder of the organization.}
\label{FIGURE1}
\end{figure}

The network model (A) in Fig.\ref{FIGURE1} represents a real organization. It is a global mujahedin (Jihad fighters) organization which was analyzed in \cite{Sag04}. The model consists of $107$ persons and 4 regional sub-networks. The sub-networks represent Central Staffs (CS), Core Arabs (CA) from the Arabian Peninsula countries and Egypt, Maghreb Arabs (MA) from the North African countries, and Southeast Asians (SA). The organization has a relatively large Gini coefficient of the nodal degree, $0.35$, and a relatively large average clustering coefficient, $0.54$, \cite{Wat98}. In economics, the Gini coefficient is a measure of inequality of income distribution or of wealth distribution. A larger Gini coefficient indicates lower equality. The values mean that the organization possesses hubs and a group structure.

The node $n_{{\rm CS10}}$ (indicated by a red circle in Fig.\ref{FIGURE1}) is a hub having relatively large nodal degree ($K(n_{{\rm CS10}})=8$). It is believed to be the founder of the organization, and said to be the covert leader who provides operational commanders in regional sub-networks with financial support in many terrorism attacks including 9/11 in 2001. His whereabouts are not known despite many efforts in investigation and capture.

The model (A) provides with the practical implication of solving the node discovery problem. But mathematical model is more suitable than the real organization to study the extensive quantitative characteristics of the method. Here, Barab\'{a}si-Albert model \cite{Bar99} with a group structure is used as a generalization of hubs and group structure in the model (A). The Barab\'{a}si-Albert model grows with preferential attachment. The probability where a newly coming node $n_{k}$ connects a link to an existing node $n_{j}$ is proportional to the nodal degree of $n_{j}$ ($p(k \rightarrow j) \propto K(n_{j})$). The occurrence frequency of the nodal degree tends to be scale-free ($f(K) \propto K^{a}$). In the Barab\'{a}si-Albert model with a group structure, every node $n_{j}$ is assigned a pre-determined group attribute $c(n_{j})$ to which it belongs. The number of groups is $G$. The probability $p(k \rightarrow j)$ is modified to eq.(\ref{preferential}). Group contrast parameter $\eta$ is introduced. Links between the groups appear less frequently as $\eta$ increases. The initial links between the groups are connected at random before growth by preferential attachment starts.
\begin{equation}
p(k \rightarrow j) \propto \left \{ \begin{array}{ll}
          \eta (G-1) K(n_{j}) & \mbox{if $c(n_{j}) = c(n_{k})$} \\
          K(n_{j}) & \mbox{otherwise}
        \end{array}
    \right ..
\label{preferential}
\end{equation}

\begin{figure}
\begin{center}
\includegraphics[scale=0.31,angle=90]{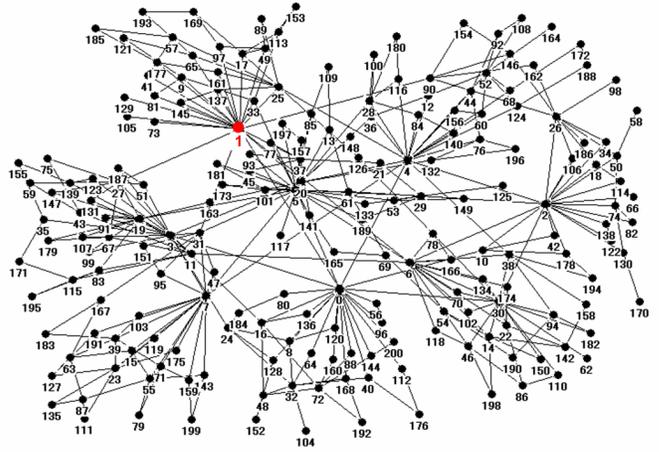}
\end{center}
\caption{Network model (C) consisting of $201$ nodes and $G=8$ groups. The group contrast parameter in eq.(\ref{preferential}) is $\eta (G-1)=400$. The node $n_{1}$, which is indicated by a red circle, is the largest hub whose nodal degree is $K(n_{1})=23$.}
\label{FIGURE3}
\end{figure}

The network models (B), (C), and (D) are examples where the number of nodes is $201$ and $G=1$, $8$, and $201$. The model (B) results in the conventional Barab\'{a}si-Albert model. The model (C) is shown in Fig.\ref{FIGURE3}. The model (D) results in Erd\"{o}s-R\'{e}nyi model \cite{Erd59} which is configured by random connection between nodes.


\subsection{Communication Model}
\label{Communication}

The dataset for performance tests is generated from the network models in \ref{Network} in the 2 steps below.

In the first step, the communication patterns $\{\delta_{i}\}$ are generated $D$ times according to the influence transmission under the true value of $\mbox{\boldmath{$\theta$}}$. A pattern includes both an initiator node $n_{j}$ and multiple responder nodes $n_{k}$. An example is $\delta_{0} = \{ n_{{\rm CS1}}, n_{{\rm CS2}}, n_{{\rm CS4}}, n_{{\rm CS5}}, n_{{\rm CS6}}, n_{{\rm CS7}}, n_{{\rm CS10}}, n_{{\rm CS11}},n_{{\rm CS12}}\}$ for the model (A) in Fig.\ref{FIGURE1}.

In the second step, the observation dataset $\{ d_{i} \}$ is generated by deleting the covert nodes in $\mbox{\boldmath{$C$}}$ from the patterns $\{\delta_{i}\}$. The example $\delta_{0}$ results in the observation $d_{0} = \delta_{0} \cap \overline{\mbox{\boldmath{$C$}}} = \{ n_{{\rm CS1}}, n_{{\rm CS2}}, n_{{\rm CS4}}, n_{{\rm CS5}}, n_{{\rm CS6}}, n_{{\rm CS7}}, n_{{\rm CS11}}, n_{{\rm CS12}}\}$ if the experimental condition is that $\mbox{\boldmath{$C$}} = \{n_{{\rm CS10}}\}$.

The covert node in $\mbox{\boldmath{$C$}}$ may appear multiple times in the communication patterns $\{ \delta_{i}\}$. The number of the target observation data to identify is given by $D_{{\rm t}} = \sum_{i=0}^{D-1} B(d_{i} \neq \delta_{i})$. The function $B(s)$ returns $1$ if the statement $s$ is true and $0$ otherwise. A few conditions are assumed in the performance evaluation in \ref{Performance} for simplicity. At first, the probability $f_{j}$ does not depend on the nodes ($f_{j} = 1/|\mbox{\boldmath{$O$}} \cup \mbox{\boldmath{$C$}}|$). Second, the value of the probability $r_{ij}$ is either 0 or 1. The number of the possible communication patterns is bounded (less than or equal to the number of nodes $N$). Finally, the influence transmission is bi-directional ($r_{jk} = r_{kj}$).

\section{Performance Evaluation}
\label{Performance}

\subsection{Performance Measure}
\label{measure}

Precision, recall, and F measure are used as a measure of the performance. In information retrieval (such as search, document classification, and query classification), the precision $p$ is used as evaluation criteria, which is the fraction of the number of relevant data to the number of the all data retrieved by search. The recall $r$ is the fraction of the number of the data retrieved by search to the number of the all relevant data. The relevant data refers to the data where $d_{i} \neq \delta_{i}$. They are given by eq.(\ref{precision}) and eq.(\ref{recall}) They are functions of the number of the retrieved data $D_{{\rm r}}$. It can take the value from 1 to $D$. The data is retrieved in the order of $d_{\sigma (1)}$, $d_{\sigma (2)}$, to $d_{\sigma (D_{{\rm r}})}$. 
\begin{equation}
p(D_{{\rm r}}) = \frac{\sum_{i=1}^{D_{{\rm r}}} B(d_{\sigma (i)} \neq \delta_{\sigma (i)})}{D_{{\rm r}}}.
\label{precision}
\end{equation}
\begin{equation}
r(D_{{\rm r}}) = \frac{\sum_{i=1}^{D_{{\rm r}}} B(d_{\sigma (i)} \neq \delta_{\sigma (i)})}{D_{{\rm t}}}.
\label{recall}
\end{equation}

The F measure $F$ is the harmonic mean of the precision and recall \cite{Kor97}. It is given by eq.(\ref{Fvalue}).
\begin{equation}
F(D_{{\rm r}}) = \frac{1}{ \frac{1}{2} (\frac{1}{p(D_{{\rm r}})} + \frac{1}{r(D_{{\rm r}})}) } = \frac{2p(D_{{\rm r}})r(D_{{\rm r}})}{p(D_{{\rm r}})+r(D_{{\rm r}})}.
\label{Fvalue}
\end{equation}

The precision, recall, and F measure range from 0 to 1. All the measures take larger values as the performance of retrieval becomes better.

\subsection{Result}
\label{Result}

The results of the performance evaluation using the test dataset in \ref{Communication} derived from the network models in \ref{Network} are demonstrated.

\begin{figure}
\begin{center}
\includegraphics[scale=0.44,angle=0]{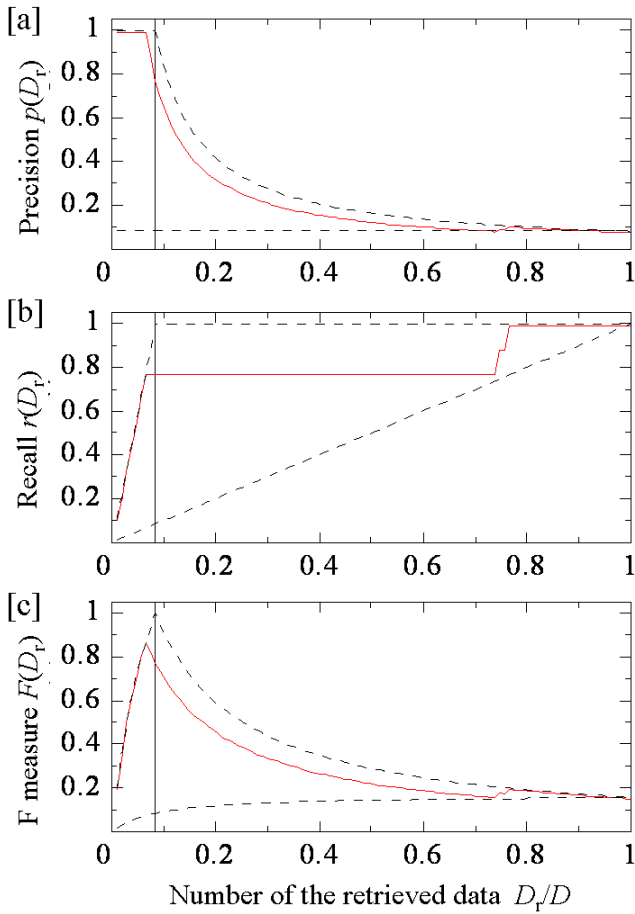}
\end{center}
\caption{Precision ($p$), recall ($r$), and F measure ($F$) in the trial where the node $n_{{\rm CS10}}$ in the model (A) is the target covert node to discover ($\mbox{\boldmath{$C$}} = \{ n_{{\rm CS10}}\}, \ |\mbox{\boldmath{$C$}}|=1$, \ $N=|\mbox{\boldmath{$O$}}|=106$). The horizontal axis is the rate of the number of the retrieved data ($D_{{\rm r}}$) to the number of the whole data ($D$). The vertical solid line indicates the position at $D_{{\rm r}}=D_{{\rm t}}$. The broken lines indicate the theoretical limit and the random retrieval limit.}
\label{FIGURE5}
\end{figure}

Let's start with the first class of the network models (real organization) and learn the implication of the method. Fig.\ref{FIGURE5} shows the precision ($p$), recall ($r$), and F measure ($F$) in the trial where the experimental condition is that the node $n_{{\rm CS10}}$ in the model (A) is the target covert node to discover ($\mbox{\boldmath{$C$}} = \{ n_{{\rm CS10}}\}, \ |\mbox{\boldmath{$C$}}|=1$, \ $N=|\mbox{\boldmath{$O$}}|=106$). The horizontal axis is the rate of the number of the retrieved data ($D_{{\rm r}}$) to the number of the whole data ($D$). The vertical solid line indicates the position at $D_{{\rm r}}=D_{{\rm t}}$. The broken lines indicate the theoretical limit (upper bound) and the random retrieval limit (lower bound). The evaluation is under the condition where the all possible communication patterns are known.

The precision, recall, and F measure are the same value of 0.78 at $D_{{\rm r}}=D_{{\rm t}}$. These are much better than those of the random retrieval ($F(D_{{\rm t}})=0.04$) and close to the theoretical limit. The method fails to discover two suspicious records $\delta_{i}$ =\{$n_{{\rm CS10}}$, $n_{{\rm CS11}}$\}, and \{$n_{{\rm CS10}}$, $n_{{\rm CS12}}$\} when $D_{{\rm r}}$ is small. This indicates that the communication with the nodes having small nodal degree ($K(n_{{\rm CS11}})=1$ and $K(n_{{\rm CS12}})=1$) does not provide much clues for node discovery. On the other hand, the most suspicious observation data $d_{\sigma(1)}$ includes all the neighbor nodes $n_{{\rm CS1}}$, $n_{{\rm CS2}}$, $n_{{\rm CS4}}$, $n_{{\rm CS5}}$, $n_{{\rm CS6}}$, $n_{{\rm CS7}}$, $n_{{\rm CS11}}$, and $n_{{\rm CS12}}$. The method succeeded in discovering most of the suspicious records and the all suspicious nodes. The investigators will decide to collect more detailed information on the communication (and a resulting collaborative activity) of the suspicious neighbor nodes. This will result in identifying, locating, and finally, capturing the covert leader ($\mbox{\boldmath{$C$}} = \{n_{{\rm CS10}}\}$) who is responsible for many terrorism attacks.

\begin{figure}
\begin{center}
\includegraphics[scale=0.30,angle=90]{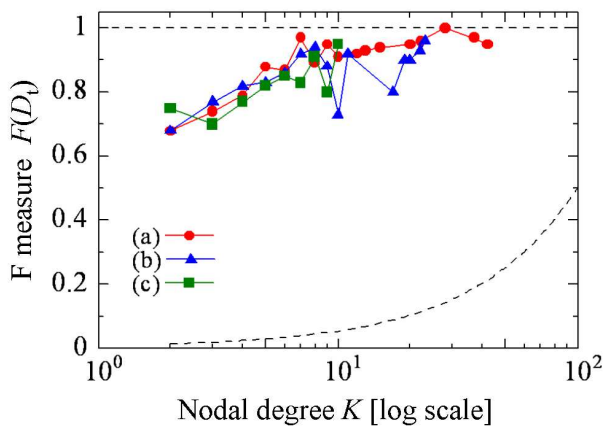}
\end{center}
\caption{F measure $F(D_{{\rm t}})$ as a function of the nodal degree $K$. Individual plots shows the F measure averaged over the trials where a node having a given nodal degree $K(n_{i})=K$ is the target covert node to discover ($|\mbox{\boldmath{$C$}}|=1$). The solid line graphs (a), (b), and (c) are for the model (B), (C), and (D) ($|\mbox{\boldmath{$O$}}|=200$). The broken lines indicate the theoretical limit and the random retrieval limit.}
\label{FIGURE6}
\end{figure}

\begin{figure}
\begin{center}
\includegraphics[scale=0.29,angle=90]{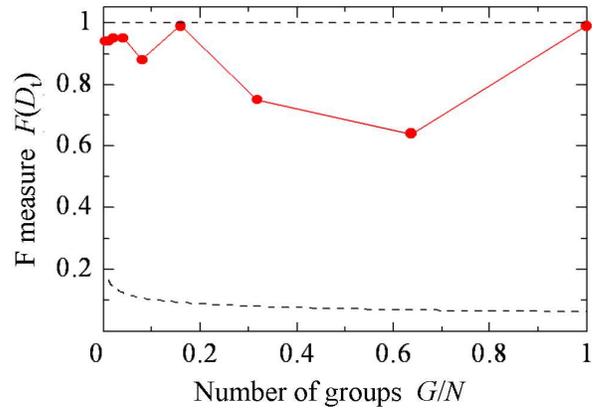}
\end{center}
\caption{F measure $F(D_{{\rm t}})$ as a function of the number of groups $G$ in the trial where the largest hub is the target covert node to discover. The number of the nodes is constant ($|\mbox{\boldmath{$O$}}|=200, \ |\mbox{\boldmath{$C$}}|=1$). The group contrast parameter is $\eta(G-1)=400$. The broken lines indicate the theoretical limit and the random retrieval limit.}
\label{FIGURE7}
\end{figure}

\begin{figure}
\begin{center}
\includegraphics[scale=0.30,angle=90]{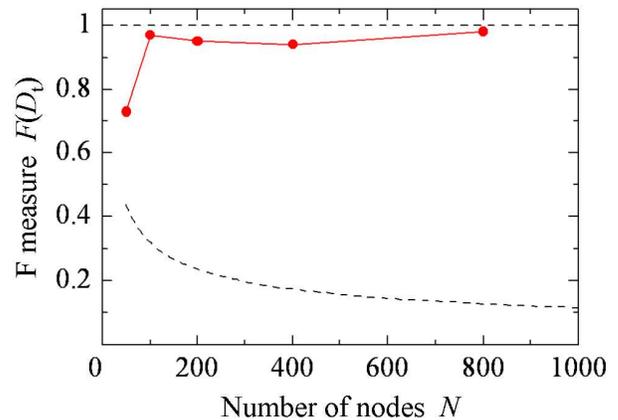}
\end{center}
\caption{F measure $F(D_{{\rm t}})$ as a function of the number of nodes $N=|\mbox{\boldmath{$O$}}|$ in the trial where the largest hub is the target covert node to discover ($|\mbox{\boldmath{$C$}}|=1$). The number of the groups is constant ($G=1$). The broken lines indicate the theoretical limit and the random retrieval limit.}
\label{FIGURE8}
\end{figure}

Let's move on to the second class of the network model (mathematical model with adjustable parameters) and study the extensive performance characteristics of the method. Fig.\ref{FIGURE6} shows the F measure $F(D_{{\rm t}})$ as a function of the nodal degree $K$. Individual plots shows the F measure averaged over the trials where the experimental condition is that a node having a given nodal degree $K(n_{i})=K$ is the target covert node to discover ($N=|\mbox{\boldmath{$O$}}|=200, \ |\mbox{\boldmath{$C$}}|=1$). The solid line graphs (a), (b), and (c) are for the model (B), (C), and (D). The broken lines indicate the theoretical limit and the random retrieval limit. The evaluation is under the condition where the all possible communication patterns are known in Fig.\ref{FIGURE6} through Fig.\ref{FIGURE8}. The resulting F measure ranges from 0.7 to 1. It does not depend on the number of groups (or topology of the network model). The performance becomes better as the nodal degree of the target covert nodes increases.

Fig.\ref{FIGURE7} shows the F measure $F(D_{{\rm t}})$ as a function of the number of groups $G$ in the trial where the largest hub is the target covert node to discover ($N=|\mbox{\boldmath{$O$}}|=200, \ |\mbox{\boldmath{$C$}}|=1$). The horizontal axis is $G/N$. The number of the nodes is constant. The group contrast parameter is fixed at $\eta(G-1)=400$ regardless of the value of $G$. The broken lines indicate the theoretical limit and the random retrieval limit. The F measure degrades down to 0.7 around $G=0.5N$. But the performance still remains much better than that of the random retrieval. The method can be applied for any value of $G$.

Fig.\ref{FIGURE8} shows the F measure $F(D_{{\rm t}})$ as a function of the number of overt nodes ($N=|\mbox{\boldmath{$O$}}|$) in the trial where the largest hub is the target covert node to discover ($|\mbox{\boldmath{$C$}}|=1$). The number of the groups is constant ($G=1$). The broken lines indicate the theoretical limit and the random retrieval limit. Except the case $N=50$, the performance remains constant. The method can be applied for large networks. Note that several hours of calculation is necessary with a standard personal computer when $N$ approaches to 1000. The size of the network is limited by the amount of calculation rather than by the accuracy obtainable from the method. 

\begin{figure}
\begin{center}
\includegraphics[scale=0.29,angle=90]{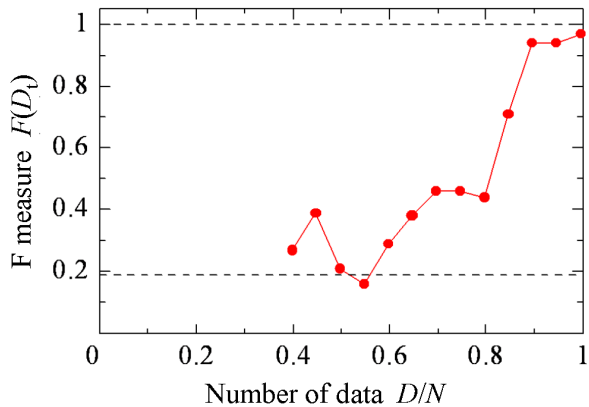}
\end{center}
\caption{F measure $F(D_{{\rm t}})$ as a function of the number of the observed data $D$ in the trial where the node $n_{0}$ in the model (B) is the target covert node to discover. The horizontal axis is $D/N$. The broken lines indicate the theoretical limit and the random retrieval limit.}
\label{FIGURE9}
\end{figure}

Fig.\ref{FIGURE9} shows the F measure $F(D_{{\rm t}})$ as a function of the number of the observed data $D$ in the trial where the node $n_{0}$ in the model (B) is the target covert node to discover ($\mbox{\boldmath{$C$}} = \{ n_{0}\}$). The horizontal axis is the ratio of $D$ to the number of the possible communication patterns ($|\mbox{\boldmath{$O$}} \cup \mbox{\boldmath{$C$}}| = N+1$) as assumed in \ref{Communication}. The ratio was $D/N=1$ in Fig.\ref{FIGURE6} through Fig.\ref{FIGURE8}. The broken lines indicate the theoretical limit and the random retrieval limit. The F measure is close to the theoretical limit, if more than 80\% of the possible communication patterns is observed. The performance is no better than that of the random retrieval if only 50\% of the possible communication patterns is observed. It is a major restriction imposed on the method that many of the possible communication patterns need to be known. Overcoming the restriction is for future work.

\section{Conclusion}
\label{Conclusion}

In this paper, I define a node discovery problem in a networked organization and present a method to solve the problem. The method infers the network behind the social interactions, applies an anomaly detection technique to the surveillance logs, and identifies the suspicious surveillance logs. The precision, recall, and F measure characteristics are close to the theoretical limit for any covert nodes in the networks of any topologies and sizes. I believe that, in the investigation of a clandestine organization \cite{Mae09}, the method aids the investigators in identifying the close associates (participants in the most suspicious surveillance record) of a covert leader or a critical conspirator.

I plan to address 3 issues for the future works. The first issue is to overcome the restriction where the performance degrades unless the ratio of the number of the observation to the number of possible communication patterns is large. The second issue is to extend the models for the social interactions. The model in this paper represents the radial influence transmission from an initiating node toward multiple responder nodes. In real networked organizations, other types of influence transmission are present such as serial (chain-shaped) influence transmission, or tree-like influence transmission. The third issue is to develop a method to solve the variants of the node discovery problem. Discovering fake nodes, or spoofing nodes are also interesting problems to uncover the malicious intentions of the organization. A fake node is the person who does not exist in the organization, but appears in the surveillance. A spoofing node is the person who belong to an organization, but appears as a different node in the surveillance.

We encounter the node discovery problem in many areas of business and social sciences \cite{Mae09b}. For example, in document analysis \cite{Tas03}, something unknown, which is not stated explicitly, can be discovered. The discovery may provide the analyst with a clue to approach the hidden intention of the author, an opinion which is about to emerge, or a sign of trends. The method will be the new basis to analyze something hidden behind the direct observation, which is beyond the scope of the conventional statistical methods and data mining expertises.

\end{document}